# Weighted Ensemble-model and Network Analysis: A method to predict fluid intelligence via naturalistic functional connectivity

Xiaobo Liu [1], Su Yang[1*]
1. School of Computing and Engineering, University of West London, London, England, UK
*means corresponding author (scottia.yang@gmail.com)

**Abstract**

*Objectives:* Functional connectivity triggered by naturalistic stimulus (e.g., movies) and machine learning techniques provide a great insight in exploring the brain functions such as fluid intelligence. However, functional connectivity are considered to be multi-layered, while traditional machine learning based on individual models not only are limited in performance, but also fail to extract multi-dimensional and multi-layered information from brain network. *Methods:* In this study, inspired by multi-layer brain network structure, we propose a new method namely Weighted Ensemble-model and Network Analysis, which combines the machine learning and graph theory for improved fluid intelligence prediction. Firstly, functional connectivity analysis and graphical theory were jointly employed. The network and graphical indices computed using the preprocessed fMRI data were then fed into auto-encoder parallelly for feature extraction to predict the fluid intelligence. In order to improve the performance, different models were automatically stacked and fused with weighted values. Finally, layers of auto-encoder were visualized to better illustrate the impacts, followed by the evaluation of the performance to justify the mechanism of brain functions. *Results:* Our proposed methods achieved best performance with 3.85 mean absolute deviation, 0.66 correlation coefficient and 0.42 R-squared coefficient, outperformed other state-of-the-art methods. It is also worth noting that, the optimization of the biological pattern extraction was automated though the auto-encoder algorithm. *Conclusion:* The proposed method not only outperforming the state-of-the-art reports, but also able to effectively capturing the common and biological pattern from functional connectivity during naturalistic movies state for potential clinical explorations.

## 1. Introduction

Human brain could be viewed as a complex network with enormous amount of locally segregated structural regions, each region dedicating to different functionalities, together they maintain global functional communications among different cognitive resources. One of the most important non-invasive approaches to measure the brain functional connectivity (FC) is the functional Magnetic Resonance Imaging (fMRI), which reflects the change of Blood Oxygen Level-Dependent (BOLD) signal [1]. As one of the major advancements in recent fMRI data analyses, functional connectivity is used to measure the temporal dependency of neuronal activation patterns in different brain regions and the communications between these regions [2]. Traditional FC analysis was based on specific experimental paradigm or resting-state; recent studies have shown that the Naturalistic Stimuli, which forms ecologically valid paradigms and approximate the real life, could improve compliance of the participants [3], hence increase the test-retest reliability [4]. Indeed, the functional connectivity with high ecological validity in naturalistic stimulus has been found more reliable than that in resting-state [5].

Many neuroimaging studies have shown that relationships between brain and cognitive functions can be established using brain measurements, cognitive measurements and statistical methods (e.g., Pearson correlation). However, statistics methods (e.g., parametric methods) tend to over-fit the data and yield a quantitatively increased certainty of the statistical estimates, while fail to generalize to novel data [6]. Furthermore, it may be impaired by high-dimensional situations (e.g., FC) [7]. On the other hand, machine learning methods with well-established processing standards, could simultaneously extract common-level patterns and leverage individual-level prediction from neuroimaging data [8]. By further integrating FC analysis into machine learning framework, a data-driven approach named connectome-based predictive modeling (CPM), could even predict individual differences in traits and behavior [9]. Coupled with the alerting score method, Rosenberg et al. found that CPM could predict sustained attention abilities using resting-state fMRI data, this finding may be applied to describe the new insight on the relationship between FC and attention [10].
Using machine learning techniques, the physiologically important representations buried in fMRI data could also be excavated and captured [11]. For example, using deep learning and fMRI, Plis et al. found that deep nets could screen out the latent relation and biological patterns from neuroimaging data [12]. These studies indicate that deep neural nets could not only be used to infer the presence of brain-behavior

(e.g., FC and human behavior) relationships and bring new representation to explain the neural mechanisms, but also can be used as the fingerprint to translate neuroimaging finding into practical utility [13]. However, traditional machine learning model based on single model was limited in model generalization and model performance [9]. Previous studies have demonstrated that the ensemble learning, proposed by Breiman et al. [14], has been integrated with bootstrap samples and multiple classifiers to improve the generalization. In addition, the overfitting issue would also be eliminated by using ensemble learning [15]. Brain networks are considered hierarchical with information processed in different layers[16]. Inspired by this, combining hierarchical structure and ensemble learning could be an effective way to improve the performance of models and extract biological information from data.

In this study, we propose a new machine learning hierarchical structure to predict the fluid intelligence, using the biological patterns extracted by measuring the functional connectivity. A new regression method based on machine learning and graph theory, namely Weighted Ensemble-model and Network Analysis (WENA) has been developed for this purpose. Compared with the traditional CPM, we used a self-supervised learning method named auto-encoder (AE) to extract no-linear and deep information from the functional connectivity and graphical theory indices based on fMRI data. In order to further improve the prediction performance, we also proposed a new method namely Weighted-Stacking (WS) which was multi-stacking-layers structure for WENA and based on the stacking structure and model fusion. The results showed that the proposed method outperforms other state-of-the-art methods, it also demonstrated the existing coherence between biological fluid intelligence and neuroimaging using this data-driven approach.

## 2. Materials and methods

### 2.1. Data acquisition

Data of 464 participants, aged from 18 to 88 years old were downloaded from the population-based sample of the Cambridge Centre for Ageing and Neuroscience (Cam-CAN, http://www.cam-can.com). The subjects without behavioral and/or neuroimaging data (fMRI or MRI) were excluded in this study, hence in total 461 controlled participants without mental illnesses and neurological disorders were included in this work. The fluid intelligence score (FIS) and other demographical information about the participants is shown in Table1.

The fMRI data were recorded while subjects watching a clip of the movie by Alfred Hitchcock named "Bang! You're Dead". According to previous neural synchronization study, the full 25-minute episode was condensed to 8 minutes [17]. Participants were instructed to watch, listen, and pay attention to the movie.

The data were collected using a 3T Siemens TIM Trio System, with a 32-channel head coil, at MRC Cognition Brain and Science Unit, Cambridge, UK. for each participant, a 3D-structural MRI was obtained using T1-weighted sequence (Generalized Auto-calibrating Partially Parallel Acquisition; repetition time = 2250 ms; echo time = 2.99ms; inversion time = 900ms; flip angle α = 9°; matrix size 256 mm × 240 mm × 19 mm; field of view = 256 mm × 240 mm × 192 mm; resolution = 1 mm isotropic; accelerated factor = 2) during the movie watching.

Table1. Demographical information of the subjects

| Total number | Age | FIS | Gender (female/male) |
| --- | --- | --- | --- |
| 461 | 54.64±18.63 | 32.97±6.30 | 231/230 |

## 2.2. Experimental Pipeline

To predict the brain fluid intelligence, we propose a novel WENA method to construct a series of model discriminative functional networks. Figure 1 illustrates the overall structure of the system. To start with, the raw fMRI data was preprocessed and the FCs (12720 FCs for each subject) from 160 regions of interest (ROIs) computed; the graphical theory indices were also obtained in parallel within this step. The indices were entered into AE module, encoded AE features and decoded AE patterns were then obtained. Finally, all features were fed into WS structures to obtain the FIS for each subject.

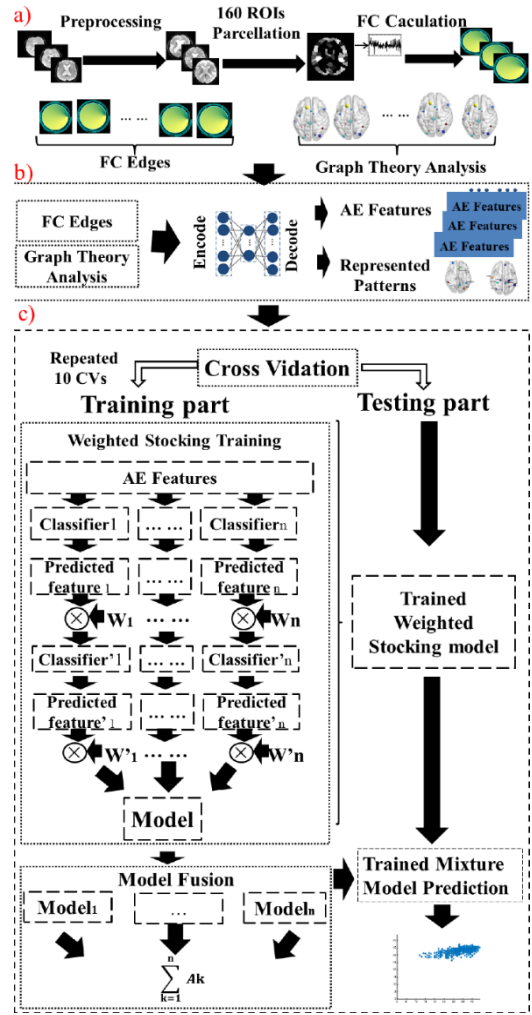

Fig.1. The overall procedure of proposed method. a) data preprocessing. b) Encode functional connectivity and graphical theory indices. The AE was used in this step to extract features and biological patterns from the network indices. c) the structure of weighted stacking fusion Model. Firstly, the features extracted from network edges and graphical theory indices were trained respectively in the first layer. In the next layer, weighted operators based on the training error caused by the last layer of the training model were added into label predicted by the last layer, and these weighted-labels were used as training features in next layers. The final predicted labels were the weighted sum of labels from different models.

## 2.3. Data preprocessing

The data preprocessing was carried out using the Data Processing Assistant for Statistical Parametric Mapping (SPM8, http://www.fil.ion. ucl.ac.uk/spm) and necessary hand-crafted MATLAB scripts (MATLAB 2018a). Initially, the first 5 volumes were discarded to reduce the impact of instability of the magnetic field. The preprocessing procedure of naturalistic fMRI included slice-timing correction, realignment, spatial normalization (3×3×3 mm$^3$) and smoothing [6-mm full-width at half maximum (FWHM)]. First, slice-timing correction were used for different signal acquisition between each slice and motion effect (6 head motion parameters). The possible nuisance signals, which included linear trend, global signal, individual mean WM and CSF signal, were removed via multiple linear regression analysis and temporal band-pass filtering (pass band 0.01-0.08 Hz). The calculation of head motion was the following formula:

$$\text{headmotion} = \frac{1}{M\text{-}1}\sqrt{|\Delta d_{x_i^1}|^2 + |\Delta d_{y_i^1}|^2 + |\Delta d_{z_i^1}|^2 + |\Delta d_{x_i^2}|^2 + |\Delta d_{y_i^2}|^2 + |\Delta d_{z_i^2}|^2} \quad (1)$$

Where M means number of time points of each subject; $d_{x_i^1}/d_{x_i^2}$, $d_{y_i^1}/d_{y_i^2}$ and $d_{z_i^1}/d_{z_i^2}$ are translations/rotations at each time point in the x, y and z, and $\Delta d_{x_i^1}$ means difference between $x_i^1$ and $x_{i-1}^1$. Furthermore, the subjects with translational motion >2.5 mm, rotation > 2.5°, mean absolute head displacement (mFD) >0.5 mm were excluded in this study. Next, the fMRI data were spatially normalized to the Montreal neurological institute (MNI) space by using Dosenbach [18]. Finally, the fMRI data were smoothed with a Gaussian kernel of 6 mm full width at half maximum (FWHM) to decrease spatial noise.

### 2.4. Functional connectivity and network property

For each participant, the whole-brain functional connectivities between all 160 brain regions were constructed pairwise from the preprocessed fMRI data according to Dosen Bash[19]. The FCs for each ROI pair, computed using the Pearson's correlation (PC), Mutual information (MI) [20] and Distance correlation (DC) [21] were calculated respectively, then further averaged over time toward the BOLD signals per subject. Once the whole-brain network was available, numerous measures could be expressed in terms of a graph. A threshold (the highest 20% of the weights) was set to sparse the constructed network. Graph theory analysis was performed on the sparse network for each subject with different FC calculation strategies. The graph theory indicesincluded the degree centrality (DC), ROI's strength (RS), local efficiency (LE) and betweenness centrality (BC). Finally, the features based on FC and graph theory indices were used for further feature representation via AE and regression.

### 2.5. Feature encoder and network pattern construction

Each subject's $N_{node} \times N_{node}$ connectivity matrices which were concatenated to give $N_{subject} \times N_{edge}$ matrix and graph theory indices which were $N_{subject} \times N_{graph\ indices}$ matrix were then entered into AE respectively (Fig. 1A). The number of epochs was 500 and the hidden nodes was set to 50 [22]. The AE, illustrated in Fig. 2, is a special type of neural network which is capable of conduct feature engineering. The vectors $x \in R$ was encoded into hidden representation $h \in R'$ by the activation function $f$:

$$h = f(Wx + b) \tag{2}$$

The hidden representation $h$ was decoded to reconstruction data $h \in R$ by the activation function:

$$r = g(W'h + b') \tag{3}$$

where W and W' are the weight matrices, b and b' represent the bias vectors, the classic *sigmoid(x)=1/(1+e$^{-x}$)* has been adopted as the activation function for $f$ and g.

Effectively a nonlinear principal components analysis (PCA) [23], the AE can be trained in a fully unsupervised manner. AE seeks the optimal parameters W, W', b and b' via gradient descent algorithm to minimizes the reconstruction error $L(x,r) = \|x - r\|^2$. In order to prevent overfitting, a weighted constraint parameter was used to regularizes $L'(x,r)$, shown in (4).

$$L'(x,r) = L(x,r) + \varepsilon \|W\|_2^2 \tag{4}$$

where $\varepsilon$ is regularization parameters.

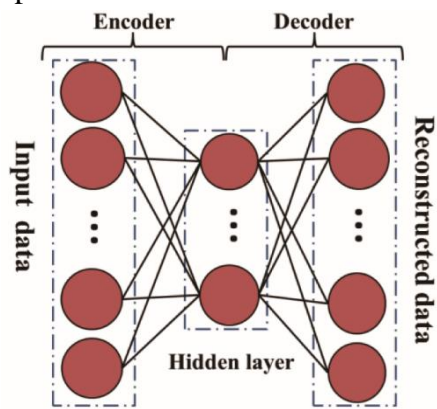

Fig.2 Autoencoder: the encoder maps input data into hidden representation, the decoder maps the encoded features to reconstruct the data

The whole-brain FC was entered into AE to extract and preserve the main information of the network according to the loss function minimum criterion [24].

## 2.6. Weighted-Ensemble models and Network Analysis Framework

All models were initially trained using different AE features, these features were extracted from network patterns and graphical indices. To prevent overfitting and the accuracy bias due to the reuse of the same data, the extracted features were split into training and test set respectively for 10-fold cross-validation. Predictive models were implemented and merged in a multi-stacking-layers approach called Weighted-Stacking (WS). On its first layer, basic regression models were used to predict FIS from neuroimaging data, weighted operators were then obtained to measure the performance of each model. The formula of weight operator *W* was shown in (5).

$$W_i = \frac{\frac{\text{Correlation Coefficient}_i}{\text{Mean Absolute Error}_i}}{\sum_{i=1}^{n} \frac{\text{Correlation Coefficient}_i}{\text{Mean Absolute Error}_i}} \quad (5)$$

Where *n* is the number of features, Correlation Coefficient refers to the correlation between real label and predicted label of each first level training model, Mean Absolute Error measures the absolute error between real label and predicted label of each first layer training model.

On the second layer, predictions from the first level models were multiplied by *W* coefficient and then stacked with other regression models. Finally, the fusion factors were set to fuse the weighted stacking models. And Fusion operator *W'* were defined in (6).

$$W'_j = \frac{\frac{\text{Correlation Coefficient}'_j}{\text{Mean Absolute Error}'_j}}{\sum_{j=1}^{m} \frac{\text{Correlation Coefficient}'_j}{\text{Mean Absolute Error}'_j}} \quad (6)$$

Where *m* is the number of regression models, "Correlation Coefficient'" indicates the correlation coefficient between real label and predicted label of each second level training model, "Mean Absolute Error" is the mean absolute error between real label and predicted label of each second layers training model.

In this study, basic regression models employed for WENA were ensemble tree regression (ETR) and ridge regression (RR). Support vector regression (SVR) with Gaussian kernel and extreme learning machine regression (ELMR) were also used to compare with the performance of WENA and test the robustness of proposed framework.

**2.7. Parameters Test**

In order to test the impact of the model parameters, in this study the stacking layers (from 2 layers to 4 layers), different FC construction methods and model fusion strategies were used to train WENA model . Also, in order to reduce the effect of other parameters on the performance, different regression models were trained via the same AE features. And we changed the parameters to be tested and fixed the others. and stacked into higher fixed stacking layers structure.

**2.8. Methods Comparison**

In this study, in order to test the performance of WENA, we compared the performance of conventional stacking models with ETR, RR, SVR and ELMR model and basic regression models. Also, features extracted via principal component analysis (PCA) and independent component analysis (ICA) were also used to trained WENA framework, the results were compared with using AE methods for feature extraction. All methods were tested in features based on three FC construction methods.

## 3. Result Evaluation

The Mean Absolute Deviation (MAE), Pearson Correlation Coefficient (R value) and R-squared Coefficient ($R^2$ value) between real value and predicted value were used to evaluate the performance of the proposed method.

### 3.1. Biological Pattern Visualization

Each AE feature was evaluated by using RelifF method [25] and the feature with the largest RelifF value was considered as the biomarker with biological significance. Pearson correlation was used to evaluate the relationship between age and AE features to extract age-related and biological patterns. The biological patterns corresponding to the chosen AE features were extracted via weight value of AE and visualized [26].

### 3.2. Results

We compared the performance of our WENA method with different weighted stacking models and FC construction methods. Table 2 illustrated that the proposed WENA achieved the best performance for fluid intelligence prediction across three functional connectivity construction methods. The performance of MI-based features obtained the highest performance with 3.85 for Mae, 0.66 for R value and 0.42 for $R^2$ value. The best FIS prediction of each network construction was shown in Fig. 3. Furthermore, conventional stacking structures and feature engineering methods were used to compare with proposed WENA method based on AE features. Table 3 showed that conventional stacking model based on SVR achieved the best performance (Mae was 4.25, R value was 0.53, $R^2$ was 0.26) with PC-network construction method and basic SVR model achieved best Mae with 4.20 (R value was 0.53, $R^2$ was 0.28). Compared with conventional feature engineering methods with MI-network-construction method, WENA achieved the performance with 4.12 for

Mae, 0.58 for R value and 0.33 for $R^2$ value for PCA methods and 4.77 for Mae, 0.32 for R value and 0.10 for $R^2$ value for ICA methods.

Stacking layers and model fusion strategies were used to test the robustness of proposed WENA. Fig 4 and Table 3 showed that proposed WENA outperformed conventional stacking models and basic regression models and was robust to network construction methods and applied stacking layer. Fig 5 showed that WENA with different regression models fusion strategies outperformed corresponding single regression models shown in Table 3.

Additionally, there was a significantly correlation found between age and FIS (R = 0.65, p < 0.001). There were also significant differences between the network AE feature and age in FC pattern (R = -0.34, p < 0.001), BC pattern (R = 0.59, p < 0.001) and LE pattern (R= 0.46, p<0.001), while there were no significant relationship found between other graph theory indices (DC and RS) and age. The most discriminative and age-related FC with network-property patterns were visualized via AE, as well as the important ROIs extracted by WENA(shown in Fig. 4 and Table 5). These results revealed that the most biological patterns extracted by WENA were sensorimotor network, cingulo-opercular network, occipital network and cerebellum network.

Table 2. The performance of weighted stack model and model fusion

| Feature | Method | MAE | R | $R^2$ |
|---|---|---|---|---|
| PC | WS- ETR | 4.21 | 0.57 | 0.31 |
|  | WS–RR | 4.07 | 0.59 | 0.33 |
|  | WS-SVR | 4.21 | 0.55 | 0.28 |
|  | WS-ELMR | 4.47 | 0.54 | 0.21 |
|  | **WENA** | **4.05** | **0.61** | **0.36** |
| MI | WS-ETR | 4.06 | 0.63 | 0.36 |
|  | WS- RR | 3.90 | 0.64 | 0.39 |
|  | WS-SVR | 4.11 | 0.60 | 0.35 |
|  | WS-ELMR | 4.43 | 0.57 | 0.24 |
|  | **WENA** | **3.85** | **0.66** | **0.42** |
| DC | WS-ETR | 4.20 | 0.56 | 0.31 |
|  | WS- RR | 4.32 | 0.56 | 0.28 |
|  | WS-SVR | 4.38 | 0.52 | 0.25 |
|  | WS-ELMR | 4.55 | 0.52 | 0.19 |
|  | **WENA** | **4.16** | **0.58** | **0.34** |

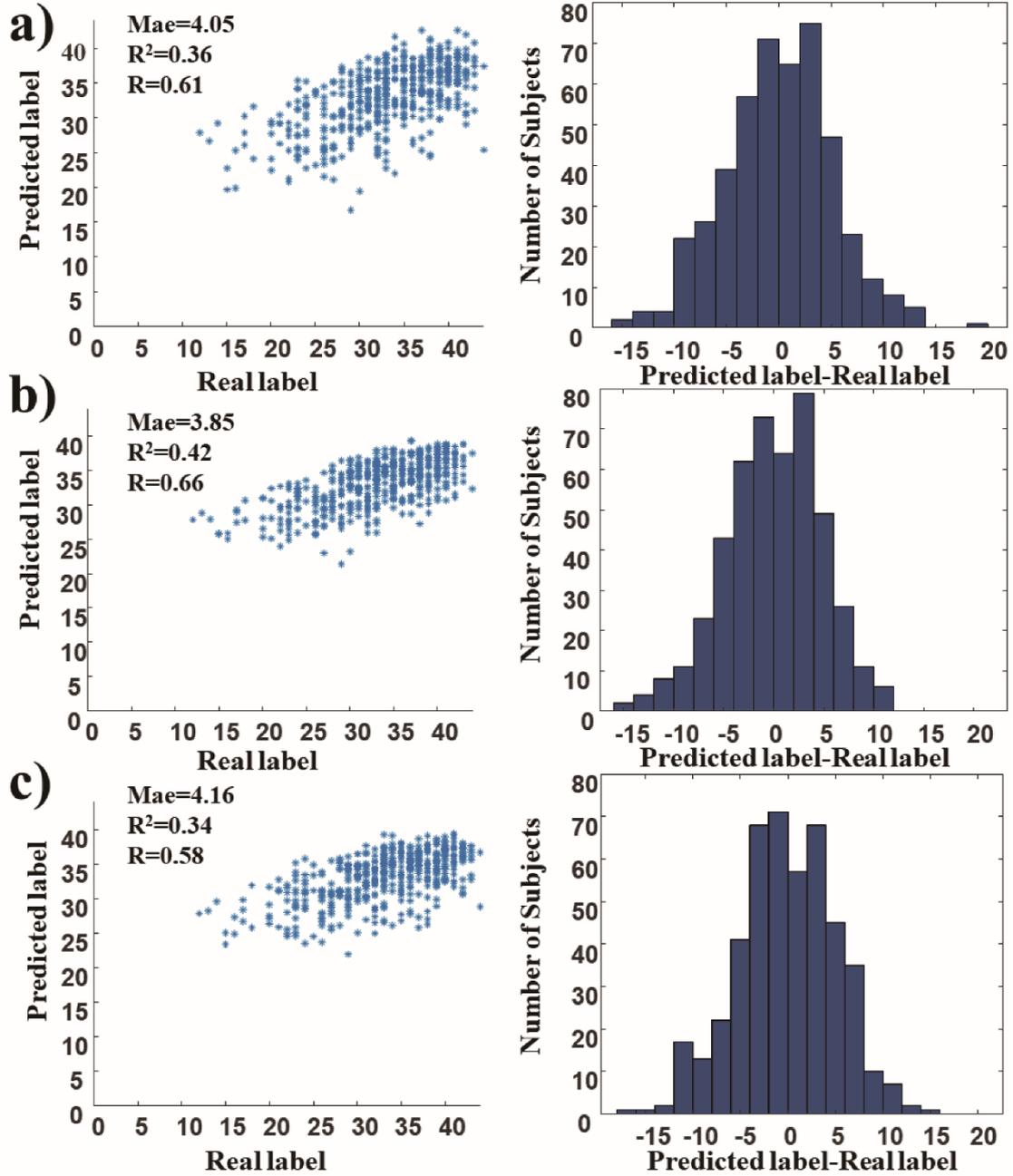

Fig.3. The best prediction performance of FIS based on different construction methods. a) the regression performance based on network based on Pearson's correlation (Mae=4.05, $R^2$=0.36. R=0.61). b) the network based on Multi information (Mae=3.85, $R^2$=0.42. R=0.66). c) the network based on Distance correlation (Mae=4.16, $R^2$=0.34. R=0.58). (Left: the performance of regression, x-coordinate represents predicted label, y-coordinate represents real label Right: the distribution of label difference, x-coordinate represents number of subjects, y-coordinate represents difference between predicted label and real label).

Table.3. The performance of conventional stacking models and single models. The performance of conventional stacking methods under different FC construction methods were obtained in order to compare the performance of WENA .

| Feature | Method | MAE | R | $R^2$ |
| --- | --- | --- | --- | --- |

| | Stacking –ETR | 4.26 | 0.53 | 0.28 |
|---|---|---|---|---|
| PC | Stacking–RR | 5.05 | 0.054 | 0.0041 |
| | **Stacking–SVR** | **4.25** | **0.53** | **0.26** |
| | Stacking–ELMR | 12.16 | 0.27 | 0.0039 |
| | Stacking–ETR | 4.20 | 0.54 | 0.29 |
| MI | Stacking–RR | 5.05 | 0.038 | 0.0042 |
| | Stacking–SVR | 4.42 | 0.50 | 0.21 |
| | Stacking–ELMR | 11.62 | 0.23 | 0.0010 |
| | Stacking–ETR | 4.25 | 0.54 | 0.29 |
| DC | Stacking–RR | 5.04 | 0.25 | 0.055 |
| | Stacking–SVR | 4.33 | 0.25 | 0.061 |
| | Stacking–ELMR | 11.98 | 0.23 | 0.0038 |
| | ETR | 4.22 | 0.54 | 0.29 |
| Basic regression | RR | 4.23 | 0.52 | 0.23 |
| Models (MI) | **SVR** | **4.20** | **0.53** | **0.28** |
| | ELMR | 4.41 | 0.49 | 0.18 |

Table.4. The performance of different feature engineering method based on MI features. The performance of conventional dimension-reduction methods under different FC construction methods were obtained in order to compare the performance of AE.

| Feature | Method | MAE | R | $R^2$ |
|---|---|---|---|---|
| | WS –ETR | 4.25 | 0.54 | 0.29 |
| | WS –RR | 4.37 | 0.55 | 0.23 |
| PCA | WS –SVR | 4.24 | 0.54 | 0.27 |
| | WS –ELMR | 4.58 | 0.52 | 0.19 |
| | **WENA** | **4.12** | **0.58** | **0.33** |
| | WS –ETR | 4.86 | 0.27 | 0.0065 |
| | WS –RR | 4.92 | 0.30 | 0.0097 |
| ICA | WS –SVR | 4.77 | 0.33 | 0.092 |
| | WS –ELMR | 5.24 | 0.25 | 0.0013 |
| | **WENA** | **4.77** | **0.32** | **0.10** |

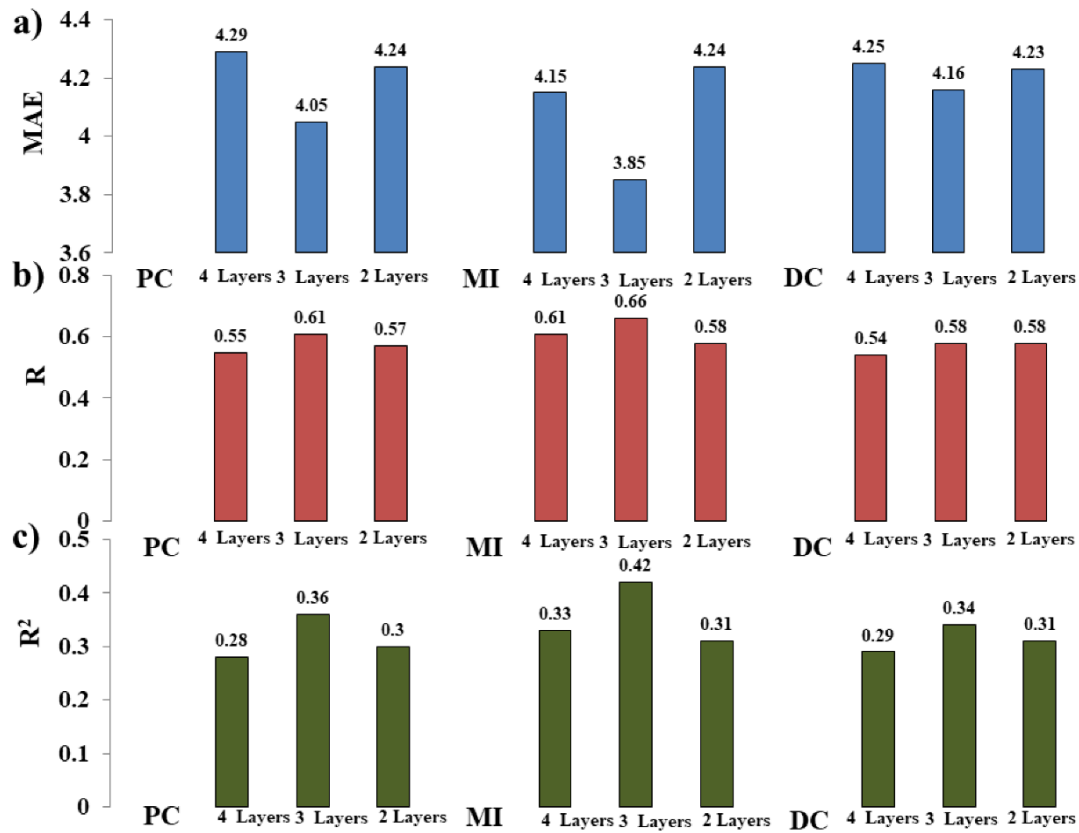

Fig. 4 The influence of stacking-layers on performance, including MAE, R value and $R^2$ value. A. Mae of WENA with different stacking-layers. B. R value of WENA with different stacking-layers. C. $R^2$ value of WENA with different stacking-layers. (X-coordinate represents MAE, R value and $R^2$ value, Y-coordinate represents number of model stacking-layers, e.g., Four- layers means this stacking model consisted of four layers).

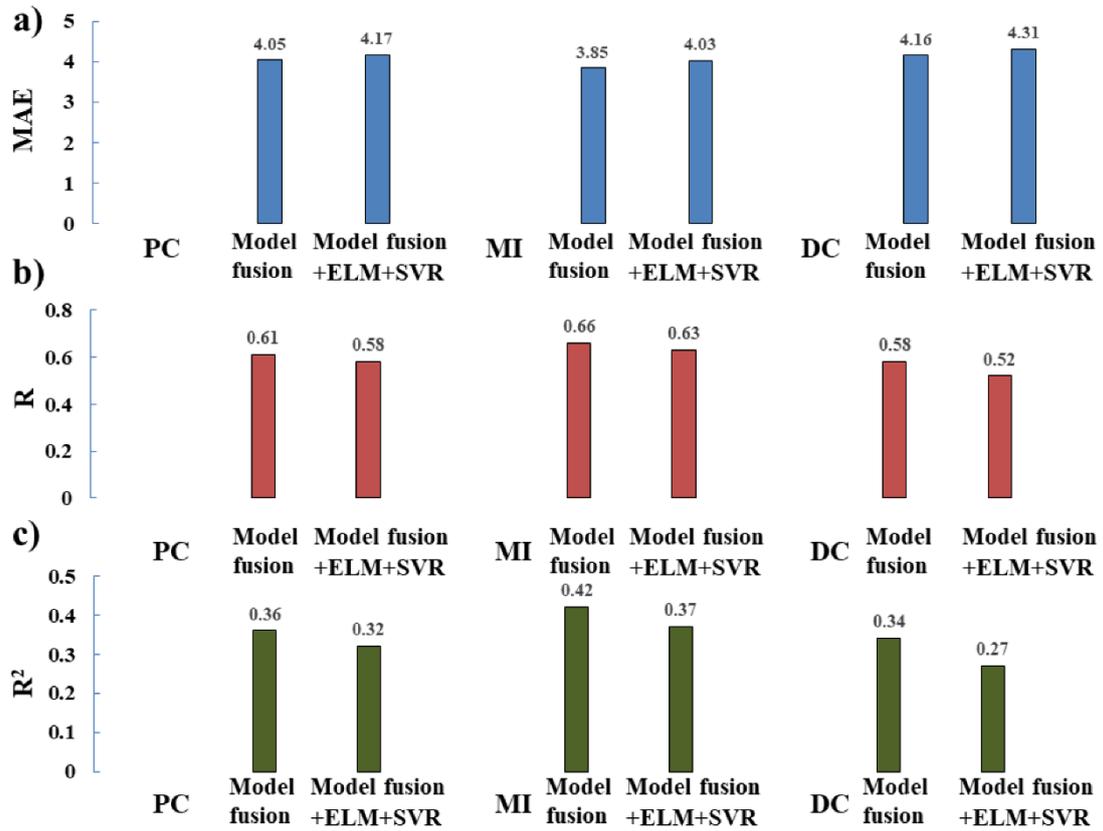

Fig.5. The influence of regression models fusion on performance of WENA with different network construction methods. A. Mae of WENA with different regression model fusion. B. R value of WENA with different classifier-choice. C. $R^2$ value of WENA with different regression model fusion. (Model-Fusion contains ETR and RR methods. X-coordinate represents MAE, R value and $R^2$ value, Y-coordinate represents different model fusion methods with different network construction methods).

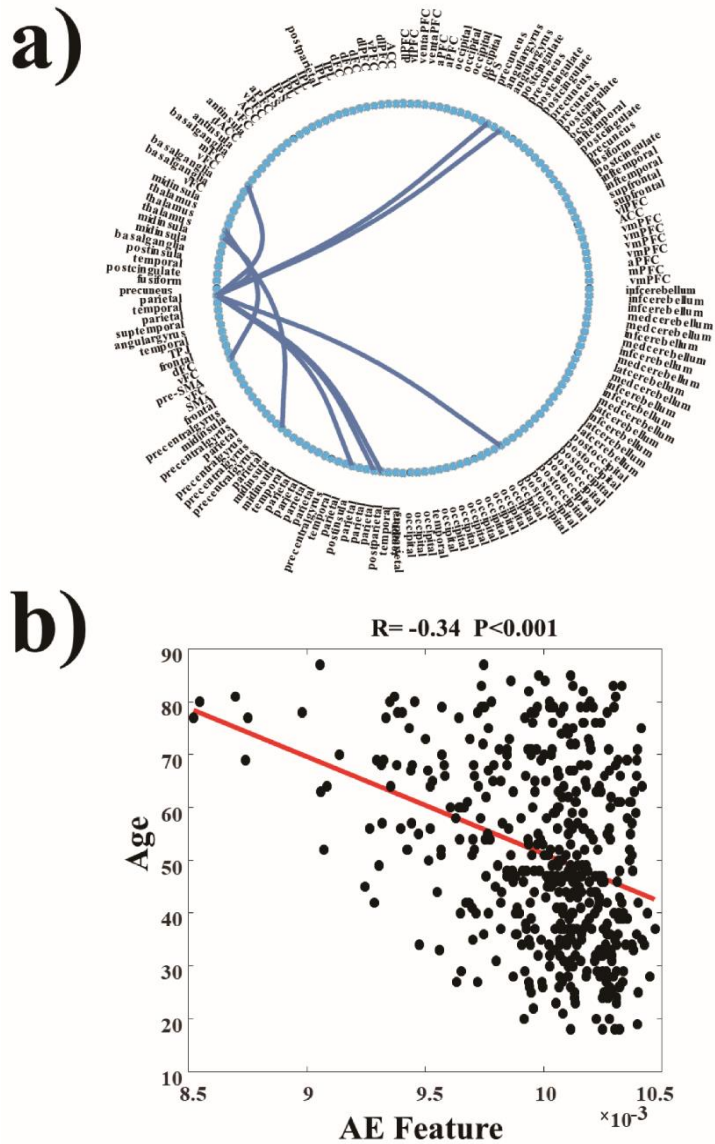

Fig. 6. The extracted network pattern via WENA. a) Network pattern. b) Pearson' correlation between age and AE feature (R = -0.34, p < 0.001. X-coordinate represents AE feature, Y-coordinate represents age).

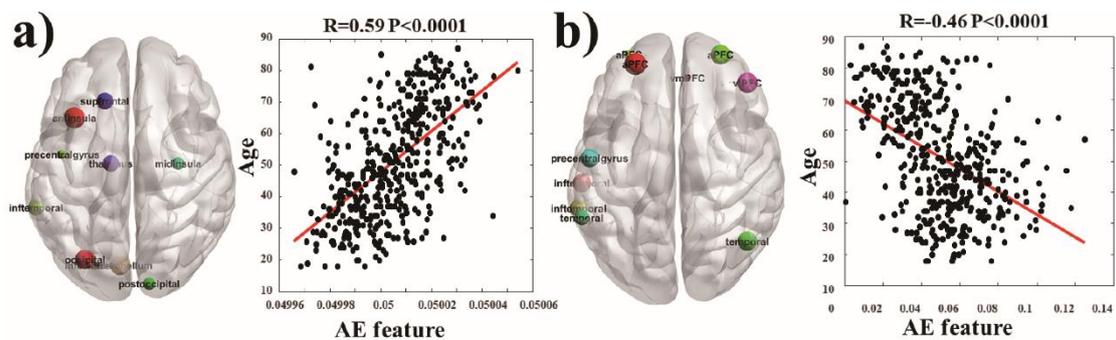

Fig.7. The extracted graphical theory indices pattern via WENA. a) BC age-related pattern (R = 0.59, p < 0.001). b) LE age-related pattern (R = -0.46, p < 0.001). (X-coordinate represents age, Y-coordinate represents AE feature.)

Table 5. the extracted important ROIs and functional networks

| network | ROIs | X | Y | Z |
|---|---|---|---|---|
| sensorimotor | Parietal | 46 | -20 | 25 |
| | precentral gyrus | 46 | -8 | 24 |
| | post insula | -30 | -28 | 9 |
| | mid insula | 32 | -12 | 2 |
| | post parietal | -41 | -31 | 48 |
| | Thalamus | -12 | -12 | 6 |
| cingulo opercular | aPFC | -25 | 51 | 27 |
| | vFC | -48 | 6 | 1 |
| | vPFC | 34 | 32 | 7 |
| | Temporal | -41 | -37 | 16 |
| occipital | post occipital | 13 | -91 | 2 |
| | Occipital | -16 | -76 | 33 |
| cerebellum | Infcerebellum | -6 | -79 | -33 |

Table 6. The state-of-the-art of fluid-intelligence score prediction.

| | Feature | MAE | R | R2 |
|---|---|---|---|---|
| [27] | fMRI | -- | 0.2~0.5 | -- |
| [28] | fMRI | -- | 0.25~0.3 | -- |
| [29] | fMRI | -- | 0.26 | -- |

## 4. Discussion

In this study, we have developed a new regression method based on machine learning and graph theory called WENA, in order to better extract the biological patterns from functional connectivity and predict the fluid intelligence. The results indicate that (a) our proposed method outperformed the state-of-the-art reports; (b) proposed method was robust under the effect of network construction methods and other parameters; (c) the patterns extracted using this method were found with interesting biologically meaning. These patterns were significantly related to age, which were found stemmed from sensorimotor network, cingulo-opercular network, occipital network and cerebellum network.

The proposed WENA structure also outperformed other traditional methods in term of performance of FIS prediction (shown in Table 2, Table 3, Table 4). Firstly, ensemble learning models (including bagging, stacking and boosting) which consisted of several single machine learning model [30], outperformed single machine learning model. Given that single machine learning algorithm was limited in model generalization and model performance [9], while performance of ensemble learning could be improved via using bootstrap replicates and simple bagging could be improved via stacking [31]. Unlike deep learning, which risks at overfitting and lacking model generalization [32], ensemble learning could integrate with bootstrap samples and multiple classifiers, which could lead to enhancement of model-generalization and reduction of model-overfitting [14, 33].

Secondly, the proposed WENA based on WS methods and model fusion outperformed traditional stacking methods (see Table3). Thirdly, The proposed method was based on self-supervised learning AE, it could extract non-linear features and principal modes from FC data across population [34]. The performance of WENA based on WS outperformed that of WENA based on principal component analysis (PCA) and independent component analysis (ICA) (see Table 4). As traditional approaches in neuroscience, PCA and ICA were both for linear features, the performance based on PCA features and ICA features were influenced by uncertain reduced dimensions [35]. By contrast, AE could represent high-layers features and abstract low-level features (e.g., cerebrospinal fluid, cortical thickness and gray matter tissue volume) from neuroimaging data, but also general latent feature representation and improve the performance [11, 36]. For example, via AE and fMRI, Suk et al. extracted nonlinear hidden features from neuroimaging data and improved diagnostic accuracy[36].

However, it should be noted that the network construction methods were used and compared in this study (shown in Table 1) and our results showed that performance of machine learning is affected by FC construction methods (shown in Table 1 and S-Table 1). WENA was robust to network construction methods for improving the performance of FIS prediction. However, the number of stacking-layers and the regression methods could affect the performance of WENA (seen in Fig. 4 and Fig. 5). In all, our results revealed that the proposed WENA model achieved the best regression accuracy on FC constructed via MI methods (Mae = 3.85, R = 0.66, $R^2$ = 0.42). Furthermore, proposed WENA was better than other conventional methods and the state-of-the-art (shown in Table 4). Also, our results revealed that proposed method was robust to parameters and could keep performance improved.

The proposed WENA methods achieved improvement in the performance of fluid-intelligence prediction from neuroimaging data, also was able to decode the biologically age-related patterns from the naturalistic fMRI data (shown in Table 3). The fluid intelligence, as a highly age-related cognitive traits, could offer objective evidence in understanding naturalistic neuroimaging data for aging problem. For example, fluid intelligence, the ability to think and solve problem under the limited knowledge situation [37], was tended to decline with aging due to reduction in executive function of prefrontal cortex [38]. In our study, FIS was negatively related to age and extracted AE features were negatively related to age ($p<0.05$). Furthermore, the functional network extracted via AE spatial filter were sensorimotor network, cingulo-opercular network, occipital network and cerebellum network. To be specific, AE feature corresponded to sensorimotor network and cerebellum network was significantly positively correlated to age, which demonstrated that compensatory may existing age-related decline in motor function [39]. The existence of the increased sensorimotor and cerebellum functional connectivity has been found in elders, supporting the increasing interactivity across network with age [40], in line with our study. Similarly, AE features corresponded to cingulo-opercular network and occipital network were significantly negatively associated with age, in line with previous studies [41]. Previous studies have also shown that sensorimotor network was associated with sensory processing and occipital network was related to visual preprocessing [41]. Additionally, cingulo-opercular network, also referred to as salience network, was decreased with age, which was the neural factor that visual processing-speed [42]. These brain functions were closely related to movie-watching experience and ageing issue as well as fluid intelligence. Therefore, these studies supported and revealed that our methods could decode biological patterns.

However, several limitations should be noted. Firstly, the WENA was unable to clearly reflect the quantitative relationship between age, functional connectivity and fluid intelligence. Secondly, robustness of proposed methods should be further tested using samples from other resources. Finally, overfitting problem in training dataset should be carefully considered, though ensemble learning could reduce it in some degree.

## 5. Conclusion

In this study, we have proposed a new method namely WENA to predict fluid intelligence and mining deep network information naturalist fMRI data, which is based on ensemble learning, FC analysis and graph theory analysis. Results indicate that the proposed method outperformed mainstream state-of-the-art methods for the problem of interest. As a deep network, once the classifier-choice and stack-level been optimized, the performance of WENA is found rather robust. Special aging-related network pattern and its properties pattern were also able to be extracted via WENA. It is found the sensorimotor, cingulo-opercular and occipital-cerebellum are the most impacting regions for the prediction of fluid intelligence. Our future work will be focusing on addressing the existing limitations of the proposed method, hence better predicting human behavior and observing the human brain states.